\documentclass{llncs}
\usepackage{makeidx}  
\usepackage{mystyle}
\usepackage{wrapfig}
\usepackage{subcaption}
\begin{document}
\frontmatter          
\pagestyle{headings}  
\addtocmark{V-IMLOP} 
\mainmatter              
\title{Anatomically Constrained Video-CT Registration via the V-IMLOP Algorithm}
\titlerunning{V-IMLOP}  
%
\author{Seth~D.~Billings\inst{1} \and Ayushi Sinha\inst{1} \and Austin Reiter\inst{1} \and Simon Leonard\inst{1} \and Masaru Ishii\inst{2} \and Gregory~D.~Hager\inst{1} \and Russell~H.~Taylor\inst{1}}
\authorrunning{Billings et al.} 
\tocauthor{Seth D. Billings, Ayushi Sinha, Austin Reiter, Simon Leonard, Masaru Ishii, Russell H. Taylor, and Gregory D. Hager}
\institute{The Johns Hopkins University, USA,
\and
Johns Hopkins Medical Institutions, USA \\
\email{asinha8@jhu.edu}}
\maketitle              

\begin{abstract}
Functional endoscopic sinus surgery (FESS) is a surgical procedure used to treat acute cases of sinusitis and other sinus diseases. FESS is fast becoming the preferred choice of treatment due to its minimally invasive nature. However, due to the limited field of view of the endoscope, surgeons rely on navigation systems to guide them within the nasal cavity. State of the art navigation systems report registration accuracy of over $1$mm, which is large compared to the size of the nasal airways. We present an anatomically constrained video-CT registration algorithm that incorporates multiple video features. Our algorithm is robust in the presence of outliers. We also test our algorithm on simulated and in-vivo data, and test its accuracy against degrading initializations.

\end{abstract}

\section{Introduction}

Sinusitis, a disorder characterized by nasal inflammation, is one of the most commonly diagnosed diseases in the United States, affecting approximately $16\%$ of the adult population annually~\cite{Slavin05}. Functional endoscopic sinus surgery (FESS) is a minimally invasive surgical procedure used to relieve symptoms of chronic sinusitis. It is estimated that around $600,000$ endoscopic interventions are performed annually in the United States~\cite{Bhattacharyya10}. The sinuses are small, composed of delicate cartilage and surrounded by critical structures, such as the carotid artery and optic nerves. Approximately $5$--$7\%$ of endoscopic sinus procedures result in complications classified as minor, and about $1\%$ result in major complications~\cite{Dalziel06}. The use of an accurate navigation system during FESS can help reduce the rate of complications, and enhance patient safety, surgical efficiency, and outcome.

The popularity of FESS and its need for enhanced navigation have resulted in several video-CT registration algorithms. Direct methods, such as that described in~\cite{Otake15}, optimize over a similarity metric to match images obtained from endoscopic video and images rendered from CT data. Tracker-based methods use optical or magnetic trackers to track the position of the endoscope relative to the patient. Methods described in~\cite{Mirota12,Leonard16} track image features and reconstruct scaled 3D points from video using structure from motion (SfM). These points are then registered to a pre-operative model shape extracted from CT. The standard algorithm for such registrations is the Iterative Closest Point (ICP) algorithm~\cite{Besl92}. ICP is a two-step algorithm, which first finds matches between two sets of points, and then computes the transformation that aligns these matches. These two steps are repeated until convergence. Several variants of ICP have been introduced, such as Trimmed ICP, which improves robustness in the presence of outliers~\cite{Chetverikov02}. ~\cite{Mirota09} presents a variant of Trimmed ICP that accounts for scale. Probability-based variants with anisotropic noise models have also been introduced. For instance, the Iterative Most Likely Point (IMLP) algorithm~\cite{Billings15} incorporates a generalized noise model into both the registration and the correspondence steps. 

However, most of these algorithms are limited by the paucity of reliable, high-accuracy video features, resulting in sparse SfM reconstructions. This can cause registration algorithms to converge to inaccurate solutions. Therefore, state of the art experimental navigation systems report registration errors of over $1$\,mm, with commercial tracker-based systems reporting errors around $2$\,mm. This hinders reliable navigation within the sinuses, where the thickness of the boundaries is generally less than $1$\,mm, going as low as $0.5$\,mm where the roof of the sinuses separates it from the brain, and $0.2$\,mm where the lateral lamella separates it from the olfactory system~\cite{Kainz89}. By comparison, CT images can have resolutions of $0.5$\,mm or less and, ideally, a navigation system should be as accurate as the underlying CT. We present the Video Iterative Most Likely Oriented Point (V-IMLOP) algorithm, which extends the IMLP framework~\cite{Billings15} by registering additional features. More specifically, while most  algorithms rely solely on 3D point sets, V-IMLOP
also uses oriented 2D contours to compute a registration.

\section{Methods}

\subsection{Video Iterative Most Likely Point (V-IMLOP)}

V-IMLOP uses two types of image features for video-CT registration: 3D point features up to scale, and 2D oriented point features representing occluding surface contours. The registration incorporates a probabilistic framework by modeling the uncertainty of these features. The uncertainty of each 3D point is modeled by a 3D anisotropic Gaussian distribution, while the position and orientation uncertainties of a point on a 2D contour are modeled by a 2D anisotropic Gaussian and von Mises distributions~\cite{Mardia00} respectively. V-IMLOP consists of two main phases, correspondence and registration. In the correspondence phase, a match for each data feature, $\x \in \X$, is computed by selecting the model point, $\y \in \Y$  that maximizes the probability of having generated $\x$.
The choice of $\y$ forms the \emph{match likelihood function} (MLF).
Assuming zero-mean uncertainty and independence of the features in each measurement, and given a 3D point feature, $\xpc$, and a current registration estimate, $\mathbf{T} = [s, \mathbf{R, t}]$, the MLF is defined as

\begin{flalign}\label{eq:3D_likelihood}
	f_{\mathrm{match\_3d}}(\xpc &| \ypc, \Sigmapc, s, \mathbf{R, t}) = \nonumber \\
    & \frac{1}{(2\pi)^{3/2}|\Sigmapc|^{1/2}} e^{-\frac{1}{2}(\ypc-s\mathbf{R}\xpc-\mathbf{t})\trans\mathbf{R\Sigmapc^{-1}R^T}(\ypc-s\mathbf{R}\xpc-\mathbf{t})},
\end{flalign}
where $\mathbf{T}$ is a similarity transform. $\ypc$ is the 3D position on the model shape that is assumed to be in correspondence with the transformed 3D data point $\mathbf{T}(\xpc) = s\mathbf{R}\xpc + \mathbf{t}$. $\Sigmapc$ is the covariance matrix of 3D positional uncertainty for the non-transformed 3D data point, and $\mathbf{R}\Sigmapc\mathbf{R^T}$ is the covariance of the transformed 3D data point. Maximizing Eq.~\ref{eq:3D_likelihood} simplifies to computing the model point, $\ypc$, that minimizes the negative log likelihood, simplified as
\begin{eqnarray}\label{eq:3D_error}
	C_{\mathrm{3d}} = \frac{1}{2} (\ypc-s\mathbf{R}\xpc-\mathbf{t})\mathbf{^T}\mathbf{R}\Sigmapc^{-1}\mathbf{R^T}(\ypc-s\mathbf{R}\xpc-\mathbf{t})
\end{eqnarray}
Next, we define the MLF for an oriented 2D contour feature, $\x_{\mathrm{2d}} = (\x_{\mathrm{2dp}}, \mathbf{\hat{\x}}_{\mathrm{2dn}})$:
\begin{flalign} \label{eq:2D_likelihood}
	f_{\mathrm{match\_2d}}(\mathbf{x}_{\mathrm{2d}} &| \mathbf{y}_{\mathrm{3d}}, \mathbf{\Sigma}_{\mathrm{2d}}, \kappa, s, \mathbf{R, t}) = \nonumber \\
    &\frac{1}{(2\pi)^2|\mathbf{\Sigma}_{\mathrm{2d}}|^{1/2}I_0(\kappa)} e^{\kappa \mathbf{\hat{y}^T}_{\mathrm{2dn}} \mathbf{\hat{x}}_{\mathrm{2dn}} - \frac{1}{2}(\mathbf{y}_{\mathrm{2dp}}-\mathbf{x}_{\mathrm{2dp}})\trans\mathbf{\Sigma^{-1}_{\mathrm{2d}}}(\mathbf{y}_{\mathrm{2dp}}-\mathbf{x}_{\mathrm{2dp}})},
\end{flalign}
where $\mathbf{\Sigma}_{\mathrm{2d}}$ is the covariance matrix of 2D positional uncertainty for $\x_{\mathrm{2dp}}$, and $\kappa$ is the concentration parameter of 2D orientational uncertainty for $\mathbf{\hat{x}}_{\mathrm{2dn}}$. 
$\y_{\mathrm{2dp}}$ is the positional component of the model point, $\ypc$, which has been projected onto the 2D image plane of the video using a perspective projection.
The normalized orientation component, $\mathbf{\hat{y}}_{\mathrm{2dn}}$, of $\ypc$ is similarly a  projection onto the video image plane, but done by orthographic projection to avoid division by zero depth since the 3D model orientations of occluding contours are parallel to the image plane. Both $\y_{\mathrm{2dp}}$ and $\mathbf{\hat{y}}_{\mathrm{2dn}}$ are scaled to convert from metric to pixel units.

As before, maximizing Eq.~\ref{eq:2D_likelihood} with respect to $\mathbf{y}_{\mathrm{3d}}$ can be reduced to minimizing a contour match error function. 
However, we must ensure that only visible model contours are projected onto the video image planes as potential matches.
To achieve this, we use the estimated camera positions to compute the  occluding contours and render the model. The z-buffers from rendering are then used to determine the subset, $\mathbf{\Psi}$, of occluding contours that are visible to each video image.
Therefore, the contour match error
for the $j$th video frame reduces to computing $\mathbf{y}_{\mathrm{2d}}$ from the set $\mathbf{\Psi}_j$ that minimizes the projected contour match error function:
\begin{eqnarray}\label{eq:2D_error}
	C_{\mathrm{2d}} &=\frac{1}{2}(\y_{\mathrm{2dp}}-\x_{\mathrm{2dp}})\trans\mathbf{\Sigma^{-1}_{\mathrm{2d}}}(\y_{\mathrm{2dp}}-\x_{\mathrm{2dp}}) + \kappa (1 - \mathbf{\hat{y}^T}_{\mathrm{2dn}} \mathbf{\hat{x}}_{\mathrm{2dn}}) \, .
\end{eqnarray} 
An upper bound on the match orientation error is also imposed to  prevent matches of widely differing orientation.

In the registration phase, we determine an updated pose for the data points by computing the similarity transform, $\mathbf{T}$, that minimizes the total match error:
\begin{eqnarray}\label{eq:match_error}
	\mathbf{T} = \argmin_{[s,\mathbf{R,t]}}\left(\sum_{i=1}^{n_{\mathrm{3d}}} C_{\mathrm{3d}i} + \sum_{j=1}^{n_{\mathrm{cam}}}\sum_{i=1}^{n_{\mathrm{ctr}j}} C_{\mathrm{2d}ji}\right),
\end{eqnarray} 
where $n_{\mathrm{3d}}$ is the number of 3D data points, $n_{\mathrm{cam}}$ is the number of video images, and $n_{\mathrm{ctr}j}$ is the number of contour features in the $j$th video image.
The correspondence and registration phases are repeated until convergence.

Outlier rejection is performed between these phases. A fraction of 3D feature pairs with highest match error are first removed, followed by chi-square tests to identify further outliers satisfying the following inequality:
\begin{eqnarray*}\label{eq:3D_chisqr}
	(\ypci-s\mathbf{R}\xpci-\mathbf{t})\mathbf{^T}\mathbf{R}(\Sigmapci + \sigmaMatchSFM\I)^{-1}\mathbf{R^T}(\ypci-s\mathbf{R}\xpci-\mathbf{t}) > \chi2inv(p,3) \, ,
\end{eqnarray*}
where $\sigmaMatchSFM$ is the average square match distance of the current 3D inliers and $\chi2inv(p,3)$ is the inverse CDF function of a chi-square distribution with 3 degrees of freedom evaluated at probability $p$~\cite{Billings15}.
Similar chi-square tests are used to reject outlying 2D contour features, with independent tests for position and orientation using the normal approximation to the von Mises distribution~\cite{Mardia00}. An upper limit on the percent of contour outliers per video frame is also enforced.



An anatomical constraint on the optimization prevents the estimated camera positions from leaving the interior feasible region of the sinus cavity. It is enforced by computing the nearest point on the mesh surface to the optical center of each estimated camera. 
If the surface normal points away from the camera, then the interior boundary has been crossed and the registration is backed up by fractional amounts of the most recent change until a valid pose is re-acquired.

\subsection{Implementation}
In this section, we explain how we obtain the data required for V-IMLOP. The 3D data points are computed using SfM on endoscopic video sequences of about $30$ frames~\cite{Leonard16}. Our initial scale estimate 
is obtained by tracking the endoscope using an electromagnetic tracker and scaling the 3D points to match the magnitude of the endoscope trajectory. Since V-IMLOP optimizes over scale, inaccuracies in this estimate do not greatly affect registration accuracy. Our optimization is constrained by user-defined upper and lower bounds on scale to ensure that an unrealistic scale is not computed in the initial iterations when misalignment of $\X$ and $\Y$ may be very large. Each patient also has a pre-operative CT, which is deformably registered to a hand-segmented template CT created from a dataset of $52$ head CTs~\cite{Avants11}.
The model shape is thereby automatically generated by deforming the template meshes to patient space~\cite{Sinha16}. 

\begin{figure}[b]
\centering
\begin{minipage}[t]{0.43\linewidth}
	\centering
	\includegraphics[width=1\linewidth]{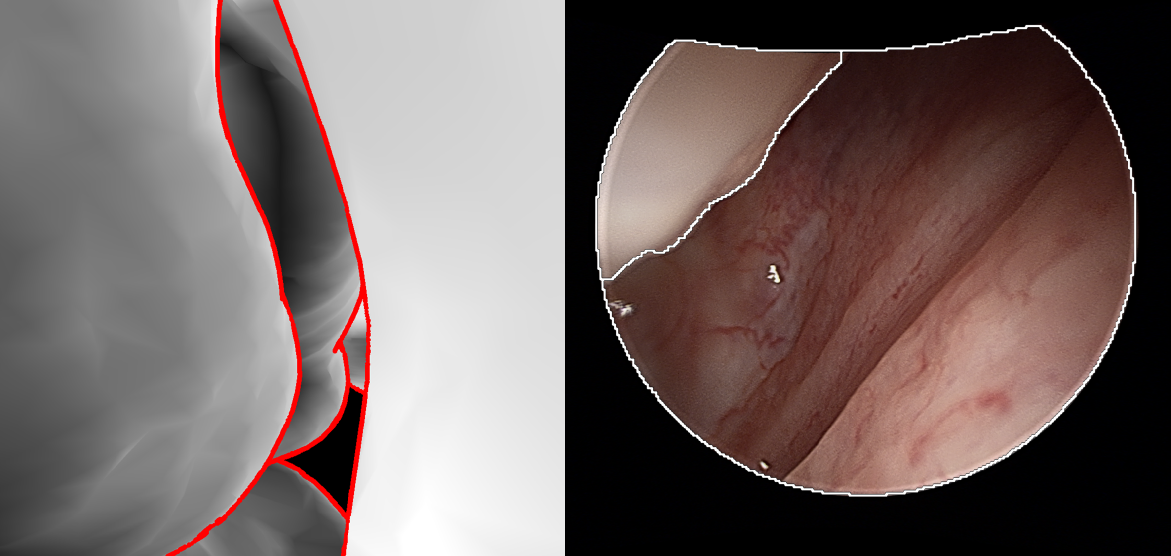} \\
    (a)
\end{minipage}
\begin{minipage}[t]{0.37\linewidth}
	\centering
	\includegraphics[width=1\linewidth]{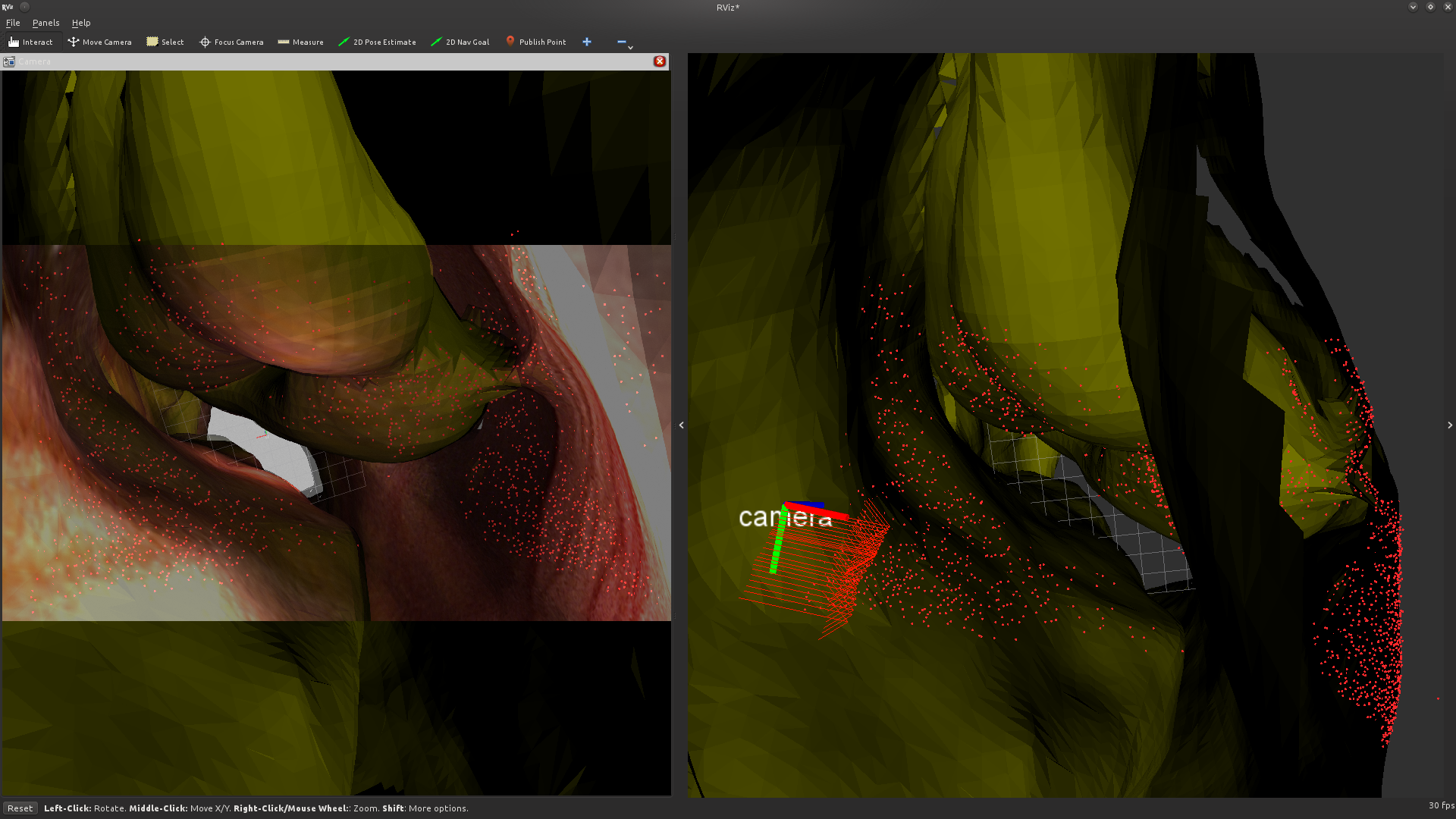} \\
    (b)
\end{minipage}
\caption[contours_overlay]
{\label{fig:contours_overlay}(a) (Left) Mesh contours in red, (right) video contours in white. (b) (Left) Overlay of CT data on the simulated image; (right) red arrows show path of the endoscope with respect to the CT, red dots show registered SfM reconstruction.}
\end{figure}

Occluding contours in video are computed once using the method described in~\cite{Arbelaez11}, because this method learns boundaries that naturally separate objects, and mimics depth-based occluding contours with high accuracy (Fig.~\ref{fig:contours_overlay}(a)). Contour normals are computed by computing gradients on smoothed video frames, and assigning to each contour point the negative gradient at that point. For the model shape, occluding contours relative to each camera pose are computed during every iteration of V-IMLOP by locating all visible edges in the triangular mesh where one face bordering the edge is oriented towards the camera, and the other away from the camera, thereby forming an occluding edge (Fig.~\ref{fig:contours_overlay}(a)). 

The measurement noise parameters ($\Sigmapc$, $\mathbf{\Sigma_{\mathrm{2d}}}$, and $\kappa$) are user defined and do not change during registration. Equal influence is granted to the 3D and 2D feature sets regardless of the feature set sizes by normalizing the user-defined 3D covariances ($\Sigmapc$) by  factor $n_{\mathrm{3d}}(1-p_t)/n_{\mathrm{2d}}$, where  $n_{\mathrm{3d}}$ and $n_{\mathrm{2d}}$ are the total number of 3D and 2D features, and $p_t$ is the initial trim ratio for 3D outliers.

\subsection{Initialization}
\label{sec:method_init}

We experiment with two approaches to initialize the registration. First, to develop a baseline, we manually set the camera pose by localizing the anatomy near the targeted field-of-view for each image. This allows us to investigate the sensitivity of V-IMLOP to the starting pose towards achieving correct convergence. Next, we relax this constraint by observing in-situ endoscopic trajectories during interventions, and isolating areas-of-interest through which the surgeon commonly inserts the endoscope. Then, we evenly sample \textit{canonical camera poses} throughout these regions, and store them in our template CT space. Through deformable registration of the template to each patient CT~\cite{Avants11}, we transform each canonical camera pose to serve as a candidate initialization from which we spawn a V-IMLOP registration process. We also slightly vary the initial scale. Finally, we select the solution yielding the minimum contour error. The residual surface error between the data points and model surface after the final transformation does not reflect failures in registration well, since SfM reconstructions are sparse, and therefore do not guarantee a unique registration. ICP and other similar methods suffer from this drawback, which causes them to often converge to solutions regardless of starting pose. Contour error, however, is a better indicator of performance. In a case where the 3D points align well but the camera pose is wrong, the projected mesh contours will not align with the video contours.



\section{Results}
\label{sec:exps}


In order to quantitatively analyze our method, we evaluated our algorithm on simulated data generated in Gazebo~\cite{Koenig04}. We used images collected from patients to texture the inside of a sinus mesh extracted from patient CT (Fig.~\ref{fig:contours_overlay}(b)). We inserted this model in a simulation with a virtual endoscope, which was navigated within the sinus cavity with physical constraints enforced by enabling collision detection in Gazebo. We computed SfM on a sequence of $30$ consecutive simulated endoscopic images, and registered the resulting $2272$ data points to the 3D mesh used for simulation (Fig.~\ref{fig:contours_overlay}(b)). Using the simulated pose of the endoscope as ground truth, we evaluated the accuracy of the method. The mean positional error for the $30$ simulated video frames was $0.5$\,mm, and orientation error was $0.49^\circ$. We also randomly sampled $900$ 3D points from the mesh visible to the virtual cameras for $6$ frames. We added random noise to the 3D points with std. dev. $0.5$\,mm and $0.3$\,mm in the parallel and orthogonal directions relative to the virtual optical axes to simulate noisy SfM data, and also to the camera poses with uniform random sampling in $[0,0.25]$\,mm and degrees of translational and rotational errors, respectively, to simulate error in the computed extrinsic parameters. Contour noise was modeled using an isotropic noise model with $\mathbf{\Sigma_{\mathrm{2d}}} = 9$\,pixel$^2$, $\kappa = 200$. The simulated data was randomly misaligned from the mesh in the interval $[2,3]$\,mm and degrees. Registration was assessed using the center points of every mesh triangle visible to the middle virtual camera frame at the ground truth pose to compute a mean TRE. Using V-IMLOP to register this data back to the mesh, we achieved a TRE of 0.2\,mm.


\begin{figure}[b!]
\centering
\begin{minipage}[t]{0.24\linewidth}
	\centering
	\includegraphics[width=1\linewidth]{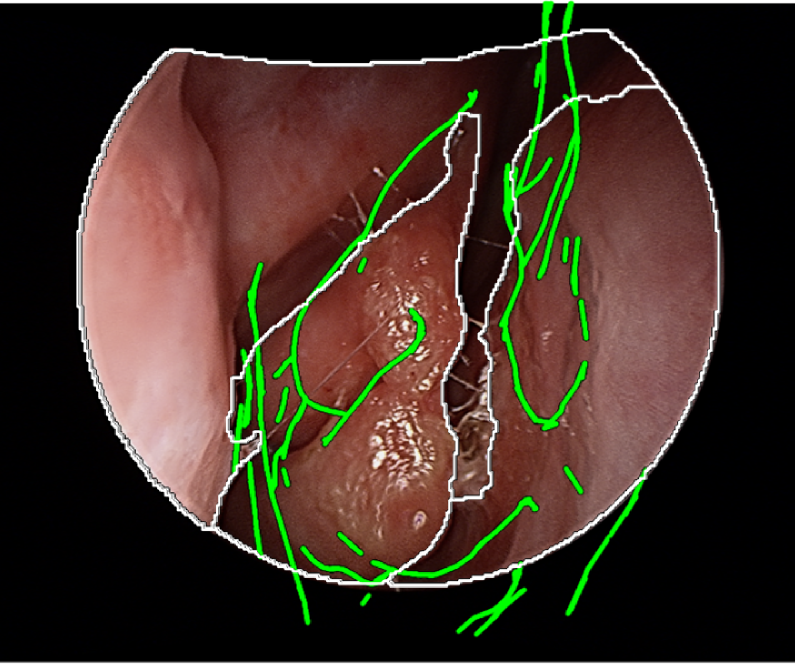} \\
    (a)
\end{minipage}
\begin{minipage}[t]{0.24\linewidth}
	\centering
	\includegraphics[width=1\linewidth]{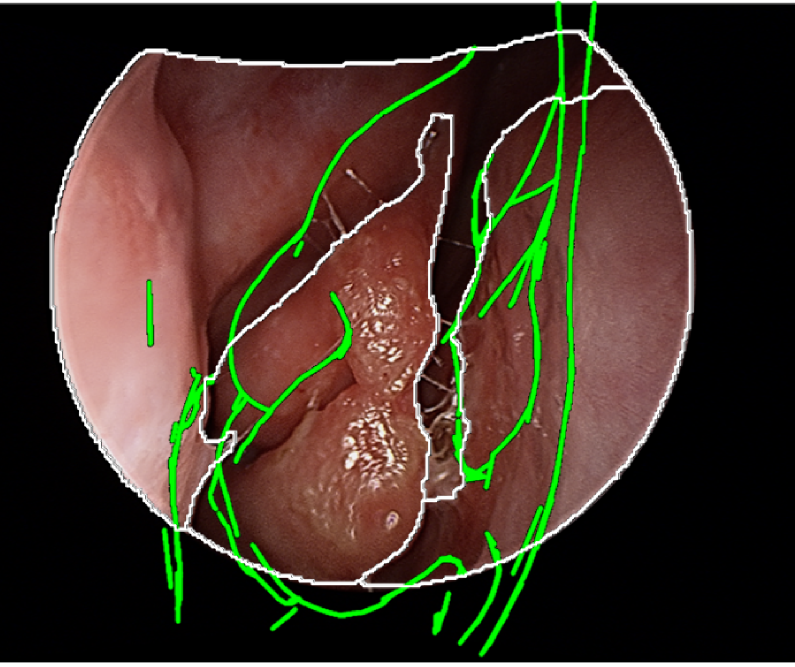} \\
    (b)
\end{minipage}
\begin{minipage}[t]{0.24\linewidth}
	\centering
	\includegraphics[width=1\linewidth]{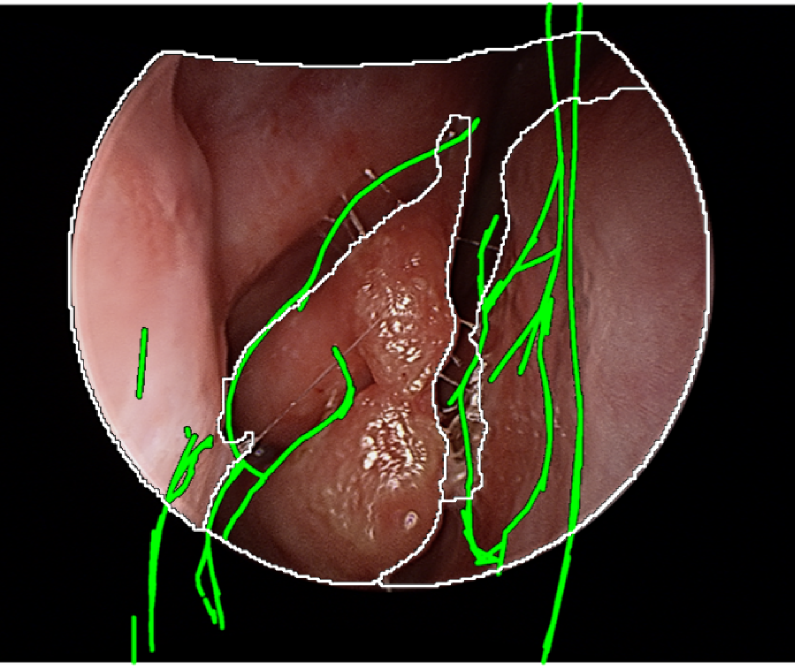} \\
    (c)
\end{minipage}
\caption[V-IMLOP]
{\label{fig:comparison}Alignment between occluding contours from CT mesh projected onto the video frame (green) and occluding contours from video (white) is better using V-IMLOP (c) than both Trimmed ICP (a) and V-IMLOP without contours (b).}
\end{figure}
\begin{figure}[t]
\centering
\begin{minipage}[t]{0.3\linewidth}
	\includegraphics[width=1\linewidth]{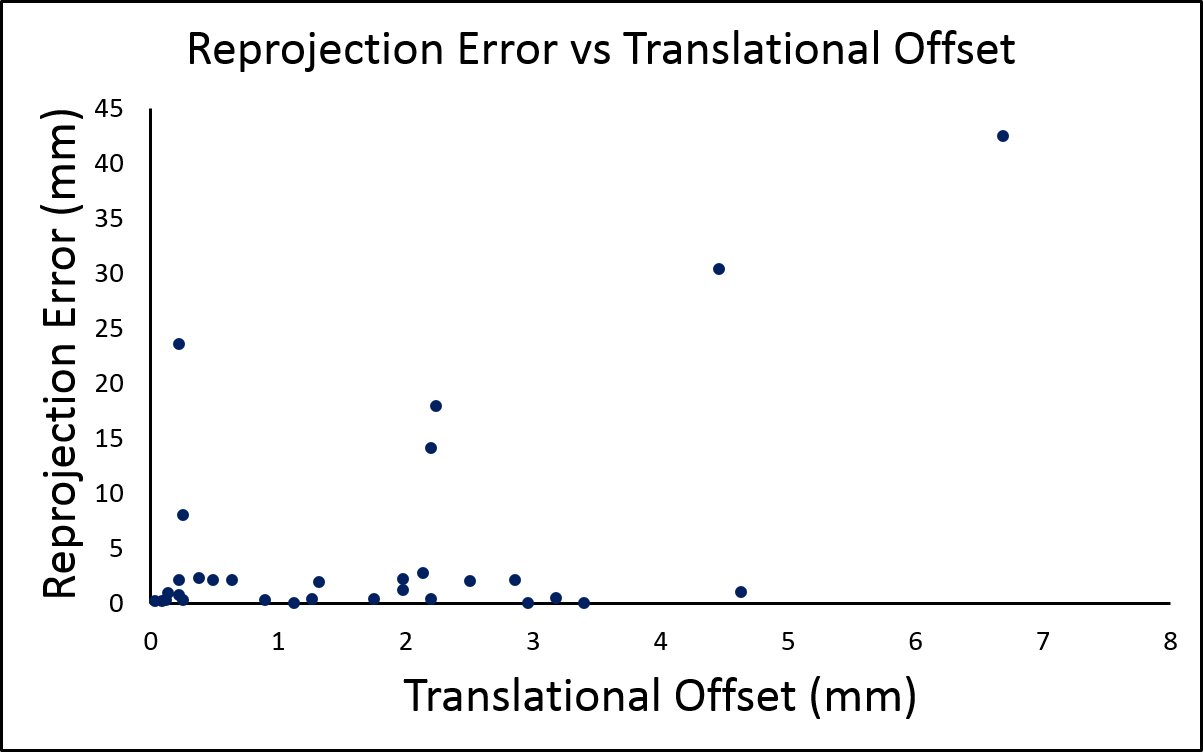}
\end{minipage}
\begin{minipage}[t]{0.3\linewidth}
	\includegraphics[width=1\linewidth]{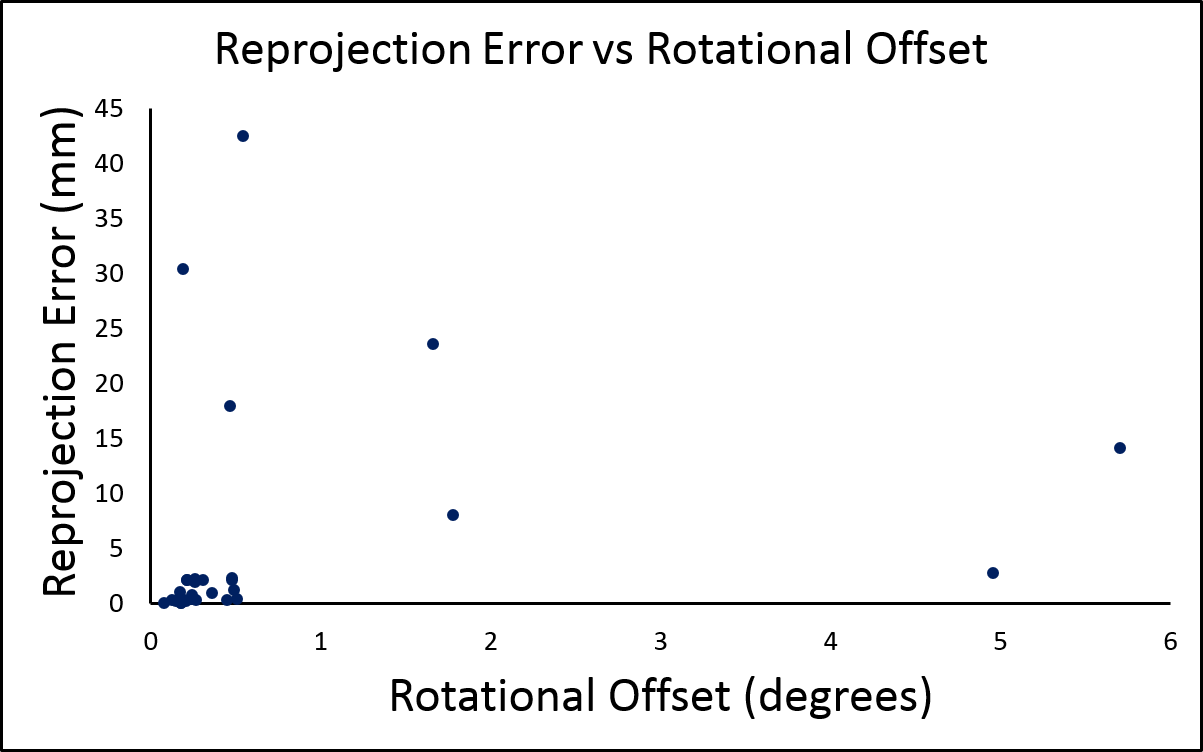}
\end{minipage}
\begin{minipage}[t]{0.25\linewidth}
	\includegraphics[width=1\linewidth]{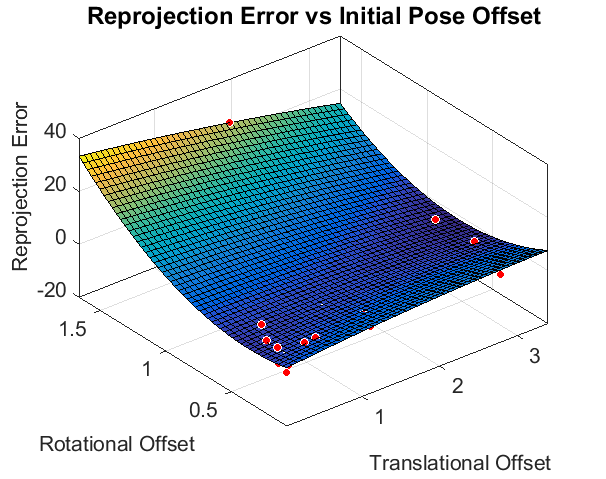}
\end{minipage}
\begin{minipage}[t]{0.3\linewidth}
	\includegraphics[width=1\linewidth]{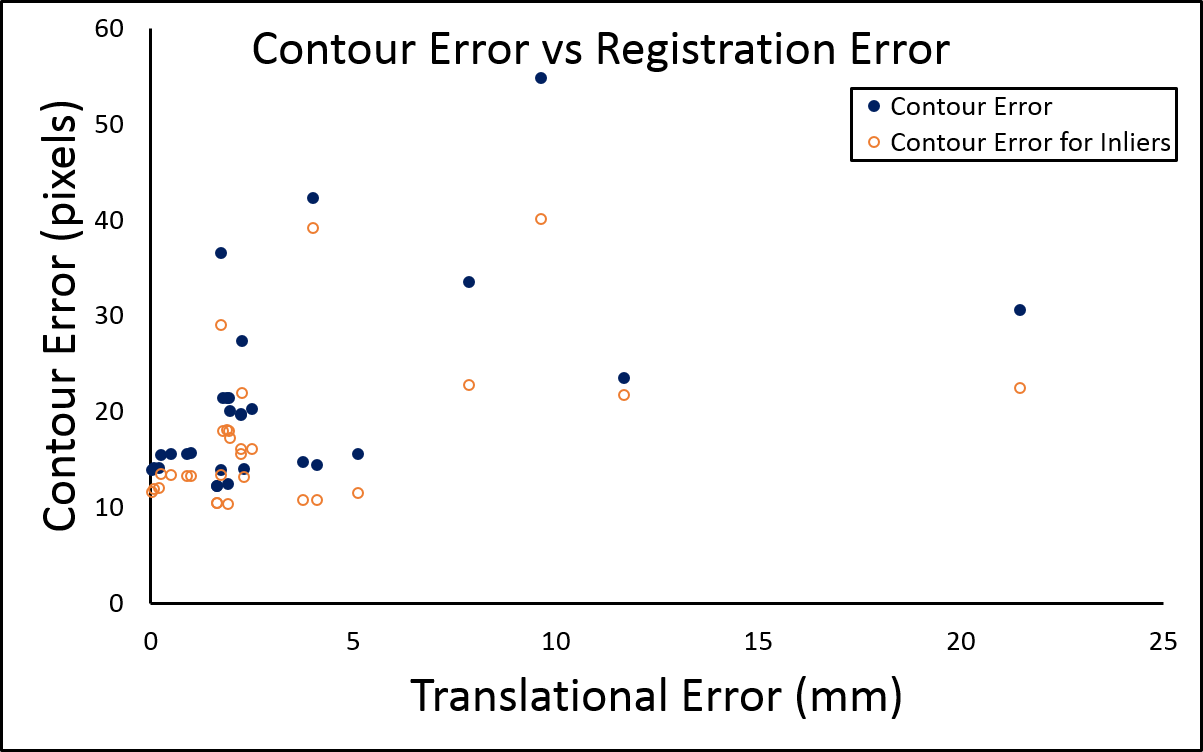}
\end{minipage}
\begin{minipage}[t]{0.3\linewidth}
	\includegraphics[width=1\linewidth]{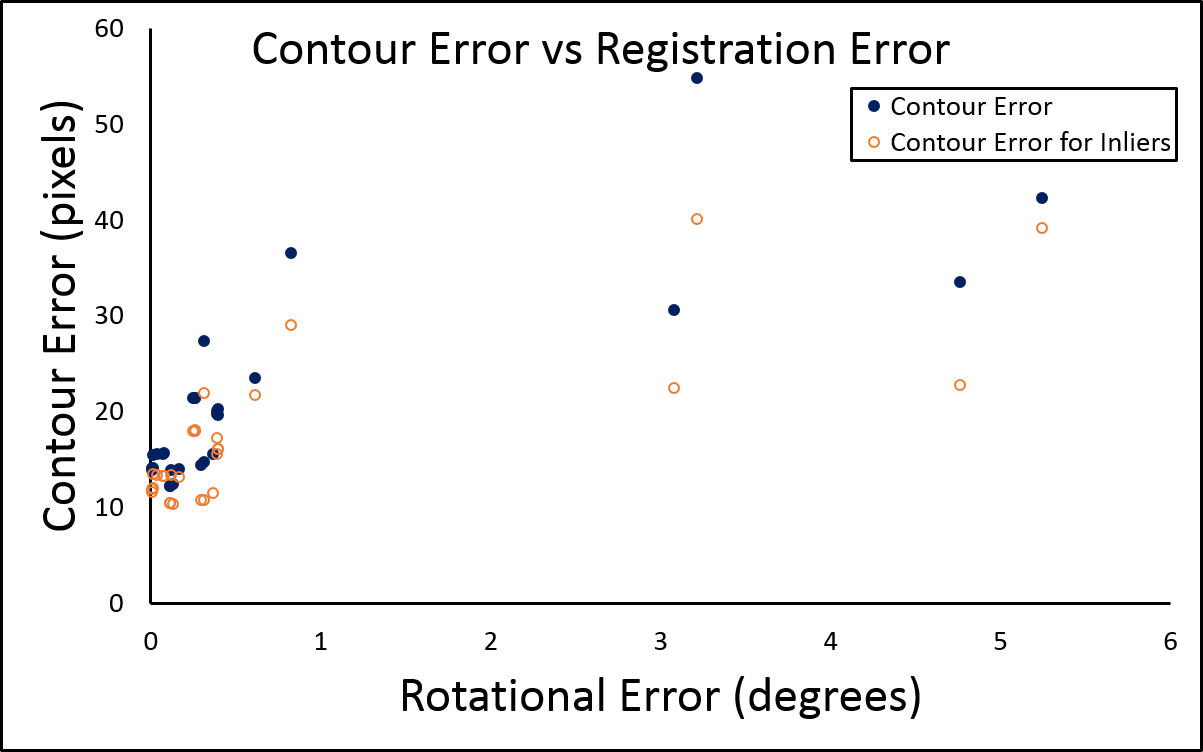}
\end{minipage}
\begin{minipage}[t]{0.25\linewidth}
	\includegraphics[width=1\linewidth]{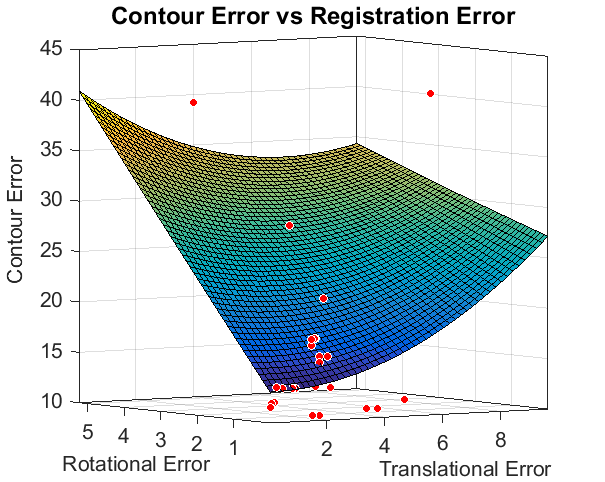}
\end{minipage}
\caption[Auto-init_error]
{\label{fig:auto_error}(Top) Registration accuracy, demonstrated through reprojection error, degrades as the initial pose is offset further from the true pose; (bottom) since contour error increases as registration error worsens, it may be used as an indicator for registration confidence.}
\end{figure}
\begin{figure}[b]
\centering
\begin{minipage}[t]{0.22\linewidth}
	\includegraphics[width=1\linewidth]{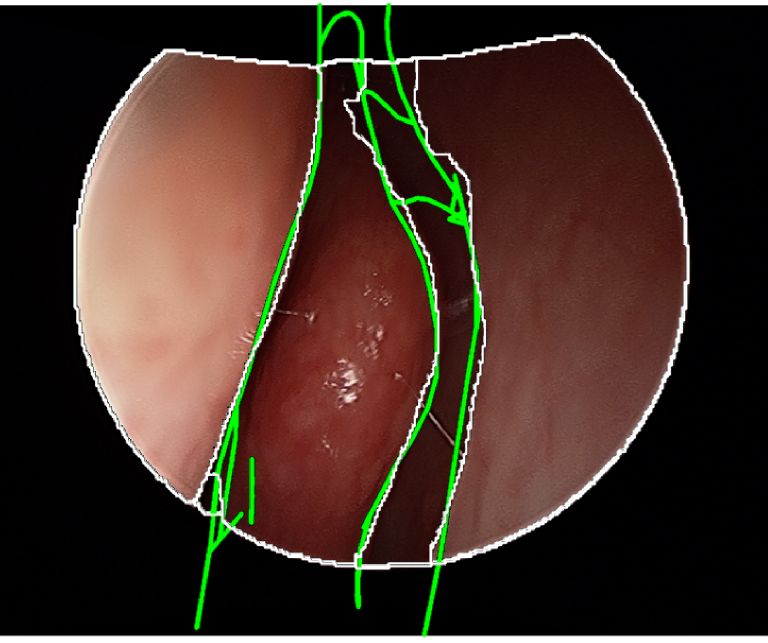}
\end{minipage}
\begin{minipage}[t]{0.22\linewidth}
	\includegraphics[width=1\linewidth]{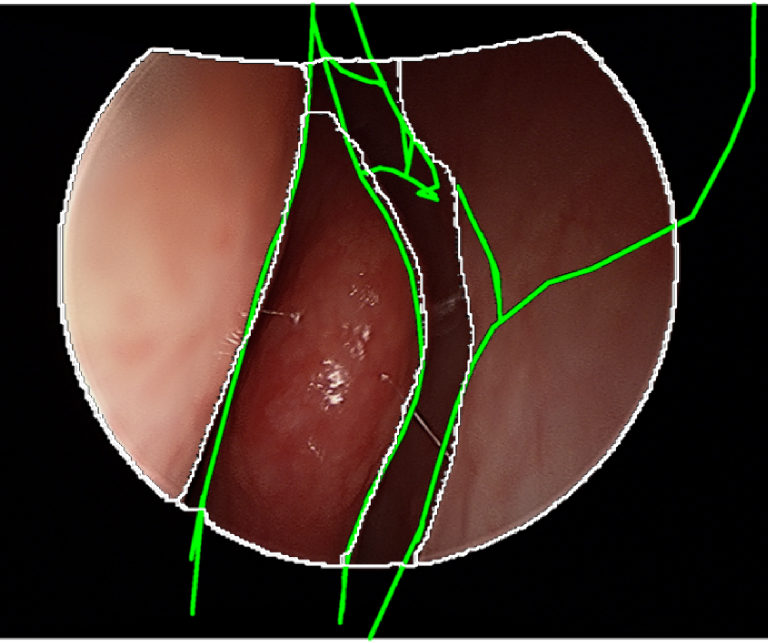}
\end{minipage}
\begin{minipage}[t]{0.22\linewidth}
	\includegraphics[width=1\linewidth]{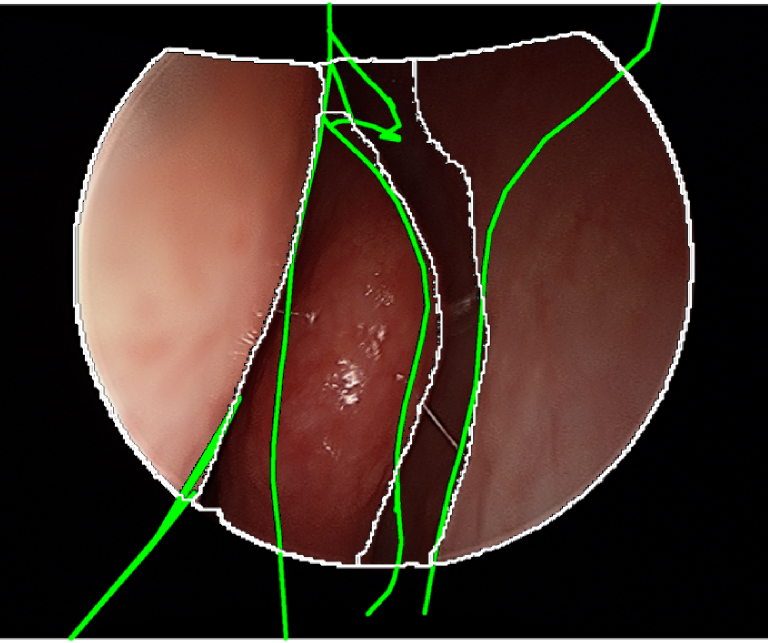}
\end{minipage}
\begin{minipage}[t]{0.22\linewidth}
	\includegraphics[width=1\linewidth]{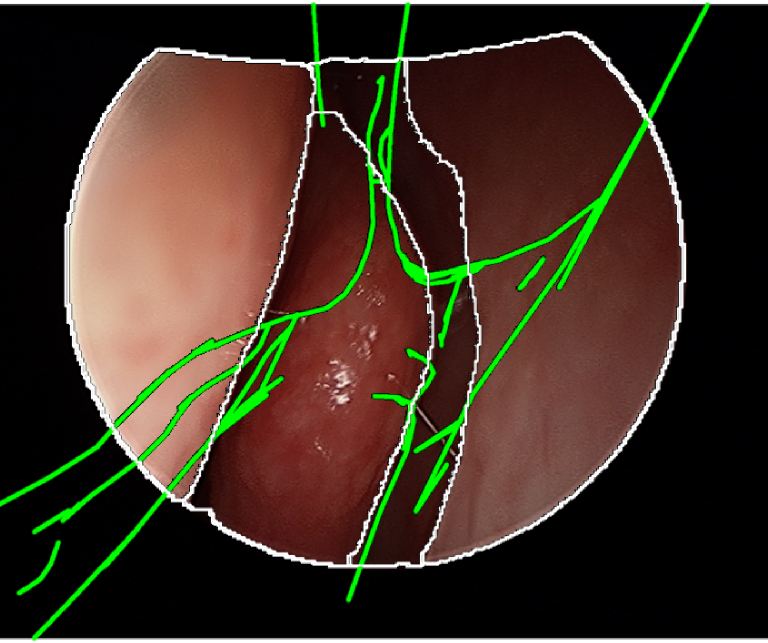}
\end{minipage}
\caption[auto-init]
{\label{fig:auto-init}Registration results using V-IMLOP with degrading initializations (left to right) show that the final registration and contour error also degrade, indicated by the alignment of model contours (green) and video contours (white).}
\end{figure}

We also tested our algorithm on in-vivo data collected from outpatients enrolled in an IRB-approved study. We tested our method with manual initializations on $12$ non-overlapping video sequences from two patients, showing differing anatomy. We used an isotropic noise model with $\Sigmapc = 0.25$\,mm$^2$, $\mathbf{\Sigma_{\mathrm{2d}}} = 1$\,pixel$^2$, $\kappa = 200$. $11$ sequences contained approximately $30$ images and $1$ sequence contained $68$, resulting in a total of $446$ images. Results from registration show that V-IMLOP produces better alignment of model contours with the corresponding video frame (Fig.~\ref{fig:comparison}). Since it is difficult to isolate a target in in-vivo data, we did not compute TRE. The mean residual error over all sequences is $0.32$\,mm, and the mean contour error is $16.64$\,pixels ($12.33$\,pixels for inliers). We also show through an analysis of perturbing our manual initializations that our approach is robust to rough pose and scale initializations, and capable of indicating \textit{failure} when a camera pose initialization is too far away from the true target anatomy. We ran this test on $31$ perturbations from $2$ sequences ($69$ images). The average residual error for the $22$ candidate initializations resulting in successful registrations was $0.25$\,mm. The right-most image in Fig.~\ref{fig:auto-init} is a failure case, and corresponds to the data point with the highest contour error in Fig.~\ref{fig:auto_error}. Therefore, we have constructed an automated initialization procedure combining empirical endoscopic trajectories with CT registration to define realistic starting poses from which registration can succeed, or return failure with confidence. 

Finally, under the guidance of a surgeon, we identify sequences with more erectile and less erectile tissue for each patient. This separation is important because structures in the sinuses that contain erectile tissue undergo regular deformation resulting in modified anatomy, and therefore, registration errors. Errors in regions of the sinuses containing more erectile tissue are $0.43$\,mm for 3D points residual error and $18.07$\,pixels ($12.92$\,pixels for inliers) contour error. Whereas, errors in regions containing less erectile tissue are $0.28$\,mm for 3D points residual error and $15.21$\,pixels ($11.74$\,pixels for inliers) contour error. Overall error is better in less erectile tissue, as expected.




\section{Conclusion and Future Work}
We present a novel approach for video-CT registration that optimizes over 3D points as well as oriented 2D contours. Our method demonstrates capability to produce sub-millimeter results even with sub-optimal initializations and in the presence of erectile tissue. We are currently working on optimizing our code, and expanding our data set to thoroughly test our method on more outpatients and surgical cases. In the future, we hope to fully automate the initialization, and further improve our method in the presence of erectile tissue by accounting for deformation. This work was funded by NIH R01-EB015530 and NSF GRFP.
%
%

\clearpage

\end{document}